\documentclass[screen,nonacm]{acmart}

\usepackage{amsmath}
\usepackage{adjustbox}
\usepackage{algorithm}
\usepackage{algorithmic}
\usepackage{diagbox}
\DeclareMathOperator*{\argmin}{argmin} 

\AtBeginDocument{%
  \providecommand\BibTeX{{%
    \normalfont B\kern-0.5em{\scshape i\kern-0.25em b}\kern-0.8em\TeX}}}

\setcopyright{acmcopyright}
\copyrightyear{2024}
\acmYear{2024}
\acmDOI{}

\begin{document}

\title{ACTER: Diverse and Actionable Counterfactual Sequences for Explaining and Diagnosing RL Policies}

\author{Jasmina Gajcin}
\email{gajcinj@tcd.ie}
\orcid{0000-0002-8731-1236}
\author{Ivana Dusparic}
\email{ivana.dusparic@tcd.ie}
\orcid{0000-0003-0621-5400}
\affiliation{%
  \institution{Trinity College Dublin}
  \city{Dublin}
  \country{Ireland}
}

\renewcommand{\shortauthors}{}

\begin{abstract}

Understanding how failure occurs and how it can be prevented in reinforcement learning (RL) is necessary to enable debugging, maintain user trust, and develop personalized policies. Counterfactual reasoning has often been used to assign blame and understand failure by searching for the closest possible world in which the failure is avoided. However, current counterfactual state explanations in RL can only explain an outcome using just the current state features and offer no actionable recourse on how a negative outcome could have been prevented. In this work, we propose ACTER (\underline{A}ctionable \underline{C}oun\underline{T}erfactual Sequences for \underline{E}xplaining \underline{R}einforcement Learning
Outcomes), an algorithm for generating counterfactual sequences that provides actionable advice on how failure can be avoided. ACTER investigates actions leading to a failure and uses the evolutionary algorithm NSGA-II to generate counterfactual sequences of actions that prevent it with minimal changes and high certainty even in  stochastic environments. Additionally, ACTER generates a set of multiple diverse counterfactual sequences that enable users to correct failure in the way that best fits their preferences. We also introduce three diversity metrics that can be used for evaluating the diversity of counterfactual sequences. We evaluate ACTER in two RL environments, with both discrete and continuous actions, and show that it can generate actionable and diverse counterfactual sequences. We conduct a user study to explore how explanations generated by ACTER help users identify and correct failure.

\end{abstract}




\maketitle

\section{Introduction}

Reinforcement learning  (RL) algorithms are being developed for many sequential decision-making tasks in various fields such as robotics, medicine, and autonomous driving \cite{li2017deep,kiran2021deep,coronato2020reinforcement}. However, these algorithms are often opaque, making their decisions difficult to understand and explain \cite{puiutta2020explainable,wells2021explainable}. The lack of algorithm transparency makes such applications less trustworthy and stands in the way of their successful integration with real-life high-stakes applications.

Of particular importance is understanding the reasons behind the failure of RL agents, as a necessary step for debugging, personalization, and developing trustworthy RL algorithms. Explanations about why failure has occurred have been shown to help recover user trust in the system \cite{tolmeijer2020taxonomy,edmonds2019tale,lyons2023explanations}. Additionally, detecting and correcting failure is necessary for debugging and correcting mistakes in the model. From the perspective of personalization, it is important to note that the definition of failure can depend on the end-user. For example, for an energy-conscious user of an autonomous vehicle, a failure is not only crashing the vehicle but also wasting fuel. Understanding why unwanted outcomes occur and how they can be avoided can help expert users know when to take over the control from the black-box model and manually control execution. Similarly, to adjust a model to their preferences, non-expert users need to understand how unwanted outcomes occurred and how they could have been prevented.

\begin{figure}
    \centering
    \includegraphics[width=1\textwidth]{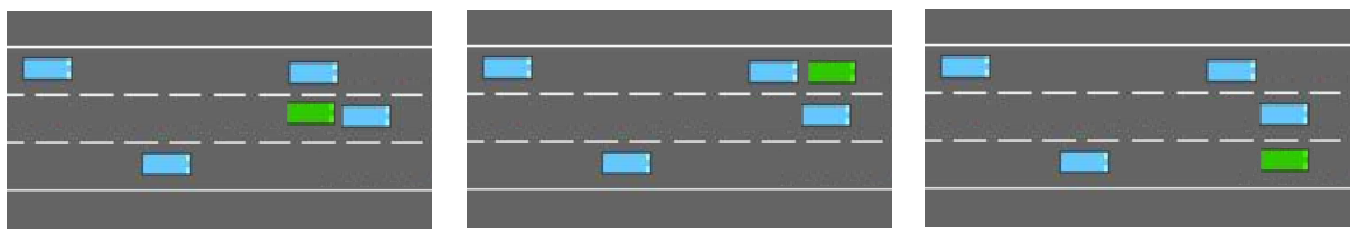}
    \caption{Lack of actionability of current counterfactual explanations. Left: ego vehicle (green) crashes with another vehicle (blue). Center and right: two possible counterfactuals stating that had the agent been in a different lane it would not have crashed. However, only the right counterfactual is actionable, as an agent could have only switched to the right lane without crashing with another car. Without examining the agent's execution history it is impossible to see how the crash could have been prevented.}
    \label{actionability}
\end{figure}

Numerous explainability techniques have been developed to understand the outcomes of RL policies. An outcome of a policy can be attributed to different factors, such as state features, previous states and actions or the agent's goals and objectives. Saliency maps \cite{greydanus2018visualizing,yang2018learn,atrey2019exploratory,itaya2021visual} and feature attribution methods \cite{puri2019explain,palsson2023unveiling} assign importance to state features based on their contribution to choosing an action. Previous work also offers methods for identifying important states in the agent's execution history, such as states in which the agent is uncertain of its actions, or states in which taking different actions leads to substantially different outcomes \cite{amir2018highlights,sequeira2020interestingness}. Furthermore, an outcome can be interpreted by inspecting the agent's reward function \cite{juozapaitis2019explainable} or its future objectives \cite{madumal2020distal,madumal2020explainable}. While the majority of these approaches focus on identifying factors responsible for a particular outcome, few works offer actionable recourse on how to change these factors to avoid a negative outcome. Counterfactual state explanations \cite{olson2021counterfactual,huber2023ganterfactual} address this shortcoming by generating an alternative state in which a negative outcome would have been avoided. Ideally, they should offer users insight into not only which features are responsible for the outcome but also how they can be changed to avoid a negative outcome. However, these explanations only consider state features and are inherently not actionable in the RL context as they do not show how those features can be changed using sequences of RL actions. For example, consider two counterfactual state explanations for a car crash in Figure \ref{actionability}. While both counterfactual states show that the crash could have been avoided had the agent changed lanes, agent could not have switched to the left lane due to another vehicle occupying it, making the counterfactual non-actionable. Without considering the agent's previous actions it is impossible to distinguish between actionable and non-actionable counterfactuals. Lack of actionability can render the counterfactual useless and decrease user trust.

In this paper, we propose ACTER (\underline{A}ctionable \underline{C}oun\underline{T}erfactual Sequences for \underline{E}xplaining \underline{R}einforcement Learning Outcomes), an approach that explores agent's past decisions to generate actionable counterfactual explanations for diagnosing and correcting undesirable outcomes in RL. Given a sequence of actions that failed, our goal is to find a \textit{counterfactual sequence} of actions, \textit {as similar as possible to the original one}, but which does not fail. Our approach can be used not only to explain and correct failure but also any other outcome of interest that users might want to better understand or prevent. We start by defining five counterfactual properties that describe the desiderata the counterfactual sequence should satisfy. Specifically, we ensure that the counterfactual sequence is as similar as possible to the original one and that it avoids the negative outcome with high certainty. To find the most suitable counterfactual according to the defined properties we use an evolutionary algorithm. Additionally, ACTER can generate a set of diverse counterfactual sequences that can prevent a negative outcome. Diverse counterfactuals are necessary for enabling personalization, as different users might have different preferences on how to best correct failure \cite{mothilal2020explaining,dandl2020multi}. For example, while all users can agree that a car crash is an unwanted outcome, some of them might prefer slowing down while others might want to change lanes to prevent it. Offering diverse counterfactuals enables users to better understand all possible ways to prevent an unwanted outcome, and choose the one that best suits their needs.

We evaluate ACTER in two environments, both with discrete and continuous action spaces and compare it to HIGHLIGHTS \cite{amir2018highlights} and interestingness elements algorithms \cite{sequeira2020interestingness}, which offer metrics for identifying the most important elements in the agent's execution. Firstly, we evaluate ACTER and the baselines on counterfactual properties to ensure that generated counterfactual sequences are similar to the original ones and successfully avoid failure even in stochastic environments. Secondly, we evaluate the diversity of generated counterfactual sequences to ensure that ACTER offers versatile ways of preventing failure. To measure diversity, we introduce three diversity metrics that can also be used to benchmark future approaches for generating counterfactual sequences. 
Finally, we evaluate ACTER in a user study. While previous work emphasizes the importance of actionable recourse \cite{karimi2021algorithmic,upadhyay2021towards,poyiadzi2020face}, user studies have reported ambiguous results, finding that counterfactuals might lead to an increase in false confidence in supervised learning tasks compared to non-actionable explanations \cite{celar2023people}. We explore this question in RL and conduct a user study to explore the effect of counterfactual sequences generated by ACTER on users' ability to diagnose and correct failure. We conduct the study using a highway driving task and compare counterfactual sequences generated by ACTER to non-actionable explanations. The main contributions of our work are as follows:

\begin{itemize}
    \item We define five desired properties for evaluating different counterfactual sequences that ensure that they are easy to obtain and avoid the negative outcome with high certainty.
    \item We propose ACTER, an algorithm for generating actionable, diverse counterfactual sequences for RL tasks 
    \item We define three benchmark metrics for evaluating the diversity of different counterfactual sequences in RL.
    \item We compare ACTER to state-of-the-art explanation techniques in their ability to robustly correct unwanted outcomes with minimal changes, the diversity of generated explanations, as well as in a user study to show the effect of counterfactual explanations on users' ability to identify and correct failure.
\end{itemize}

The full code for ACTER, baselines, and evaluation is available at: \href{https://github.com/jas97/temporalSimCFs/tree/acter}{https://github.com/jas97/temporalSimCFs/tree/acter}.

\section{Related Work}

Understanding why an unwanted outcome occurred is of utmost importance for developing, debugging and personalizing RL policies. An unwanted outcome can have many definitions, such as an unwanted action (e.g. sudden breaking) or a function of state features (e.g high speed) that a user would like to avoid. Various factors, such as state features, actions, agent's goals or objectives, can be used to explain an unwanted outcome. 

The majority of current research in explainable RL uses current state features to explain an outcome. For example, saliency maps identify parts of images that influenced the agent's decision in visual tasks \cite{greydanus2018visualizing,yang2018learn,atrey2019exploratory,itaya2021visual}. Similarly, feature attribution methods assign importance to individual features based on their contribution to action choice in a state \cite{puri2019explain,palsson2023unveiling}. Previous actions and states can also explain the agent's outcome. EDGE \cite{guo2021edge} provides global explanations that identify specific states in the episode that contributed to its outcome. EDGE trains an additional self-interpretable model to approximate the black-box policy and extract explanations for visual tasks. HIGHLIGHTS \cite{amir2018highlights} identifies critical states as those where taking different actions can lead to substantially different outcomes.
Similarly, ``interestingness metrics'' can be used to identify important states as those that are (in)frequently visited, in which the agent is (un)certain of its decision, or those where the agent's estimate of Q function reaches a local extremum \cite{sequeira2020interestingness}. The agent's goal and objectives have been used to interpret outcomes. \citet{juozapaitis2019explainable} propose a reward decomposition approach that can interpret a decision by identifying which reward components contributed to it. \citet{yau2020did} explains the agent's decision by providing the intended outcome as a motivation for the agent's action. Similarly, \citet{madumal2020distal,madumal2020explainable} envision a causal chain of future events to explain a decision. While these approaches can attribute the outcome to a factor in an agent's execution, they do not provide an actionable way to change them to avoid negative outcomes. In contrast, ACTER provides actionable advice on how to change an agent's past actions and prevent unwanted outcomes.

Understanding how an outcome could have been prevented often requires counterfactual reasoning \cite{byrne2019counterfactuals}. Counterfactual state explanations employ this reasoning to explain why a certain action was chosen by showing an alternative state where a different action would be taken. \citet{olson2021counterfactual} propose an approach based on generative modeling and generate realistic counterfactuals for Atari environments. Similarly, GANterfactual-RL is a generative model-agnostic approach that uses domain transfer architecture to generate counterfactual states \cite{huber2023ganterfactual}. Domains are determined by actions and contain states in which a specific action would be chosen by the agent. Understanding what is required for the action to change requires translating the state from its original to an alternative domain. While these explanations show what the state features should look like to prevent a negative outcome, they do not offer actionable recourse in terms of RL actions on how to actualize these changes. In contrast, our work provides actionable recourse as sequences of actions that can be used to understand and prevent failure in RL tasks.

Actionable recourse has been explored in previous RL work.
RACCER offers actionable advice on how to change features to elicit an action change \cite{gajcin2023raccer}. However, RACCER generates the counterfactual sequence by exploring the agent's future goals and not its previous behavior. In contrast, ACTER explores the agent's past and proposes actionable changes to prevent a negative outcome. Previous work also offers algorithms for finding counterfactual sequences of actions based on causal inference \cite{tsirtsis2021counterfactual,tsirtsis2023finding}. However, these algorithms are limited to discrete actions and do not offer diverse explanations. Additionally, they can only define failure when it is represented by a low reward and require a Lipschitz continuity assumption for the reward function in continuous state spaces \cite{tsirtsis2023finding}. In contrast, ACTER applies to tasks with both discrete and continuous action spaces, allows a wide range of failure definitions, and generates a set of diverse counterfactual sequences. While previous work \cite{tsirtsis2023finding,tsirtsis2021counterfactual} explores algorithmic challenges of generating counterfactual sequences, we focus on defining criteria that make these sequences useful for the users.

\section{ACTER}

In this section, we propose ACTER, our approach to explaining RL policies. ACTER can be used for diagnosing and correcting failure in the broadest context, either defined as an unwanted action or a function of state features. ACTER is model-agnostic and can be used to explain the behavior of any RL policy. In the remainder of this section, we define the problem of counterfactual sequence generation (Section \ref{Problem Definition}). We identify the desiderata a counterfactual sequence should satisfy in Section \ref{Counterfactual Properties} and provide an algorithm for finding the optimal counterfactual sequence in Section \ref{Algorithm}.

\subsection{Definitions}
\label{Problem Definition}

\subsubsection{Failure Function}

We define failure as a binary function over states $F(s): S \rightarrow [0, 1]$:

\begin{equation}
    F(s) = \begin{cases}
        1 & \text{if failure occurs in state s} \\
        0 & \text{otherwise}
    \end{cases}
\end{equation}

We assume $F(s)$ is defined beforehand. This can be done by the developer who would like to debug unwanted behavior in a policy, such as an autonomous car crash. Failure function can also be defined by a non-expert user trying to obtain a personalized policy by correcting what they perceive as negative outcomes, such as a car driving too fast for their preferences. 
Similarly, failure states can also be automatically inferred by isolating states that achieve local minima in the reward function. We leave different interpretations of failure for future work, and in this work consider more obvious failure cases, such as a car crash in a highway driving environment. 

\subsubsection{Counterfactual Sequence}

Given a failure function $F(s)$ and a state $x_n$ where $F(x_n) = 1$ we explore the sequence of factual actions $A = [a_{n-k}, \dots, a_{a_{n-1}}]$ starting in $x_{n-k}$ and leading to $x_{n}$. We limit our search to previous $k$ actions, where horizon $k$ depends on the specific task. To explain a car crash, we might focus only on a small number of previous steps, while to diagnose failure in a medical scenario we might need to look back months or even years. 
ACTER searches for a counterfactual sequence of actions $A^* = [a^*_{n-k}, \dots, a^*_{n-1}]$ that starts in $x_{n-k}$ and successfully avoids failure. 

\subsubsection{Stochastic Environment Configurations}

Counterfactual thinking requires rewriting the past. While we can identify why a car crash occurred and how it could have been prevented in a specific scenario, due to the stochasticity of the world, it is unlikely that the exact same conditions will be repeated. Counterfactual sequences should be useful to users not only in singular scenarios but in a wider range of similar situations. Thus, we need to distinguish between different types of stochastic conditions. Firstly, we define $\mathcal{P}$ as the set of stochastic processes $P_i$ outside of the agent's control:

\begin{equation}
    \mathcal{P} = \cup_{i\in I}{P^i}
\end{equation}

For example, these can describe weather conditions or the behavior of other vehicles in an autonomous driving scenario. We define a \textit{stochastic configuration} $P_T$ for a trajectory $T$ as an instantiation of all stochastic processes along $T$:

\begin{equation}
    P_T = \cup_{i \in I} P^i_T
\end{equation}

Intuitively, $P_T$ gathers the specific values that the stochastic processes took along the trajectory $T$. These can be specific weather conditions and temperature on the days of the trajectory or the specific behavior of all other cars on the road in an autonomous driving scenario. Knowing $P_T$ for any trajectory would allow one to effectively look into the past and repeat the same trajectory with all stochastic processes fixed. In this work, we assume access to RL environment, meaning we can observe the stochastic processes $P_T$ that shape the environment. However, if the full environment is not present, it can be approximated from an environment model, constructed using simulations or offline data. We leave the offline scenario for future work and explore its further implications in Section \ref{Future Work}.

\subsection{Counterfactual Properties}
\label{Counterfactual Properties}

While there are often multiple ways to prevent a failure, only some of them might be useful for the user. For example, while technically correct, telling the user that the car would not have crashed if it had never moved does not help them prevent this issue in the future. In this section, we introduce five counterfactual properties that ensure counterfactual sequences are actionable, require minimal effort to change actions, and avoid failure even in stochastic environments.  

\subsubsection{Validity}

Validity ensures that taking counterfactual actions avoids failure. For a state $x_n$ where the failure occurred ($F(x_n) = 1$), we observe trajectory $T = [(x_{n-k}, a_{n-k}), \dots, (x_{n-1}, a_{n-1})]$ with a stochastic configuration $P_T$ which leads to failure. 
Validity for a counterfactual sequence of actions $A^* = [a^*_{n-k}, \dots, a^*_{n-1}]$ is defined as:

\begin{equation}
    V(A^*) = F(A^*(x_{n-k})|_{P_T}) = 0 
\end{equation}

where $A^*(x_{n-k})|_{P_T}$ is the state reached by following $A^*$ from $x_{n-k}$ under the stochastic configuration $P_T$. This way, we ensure that in the exact same episode with the same stochastic properties, choosing $A^*$ would have prevented failure.

\begin{figure}
    \centering
    \includegraphics[width=\textwidth]{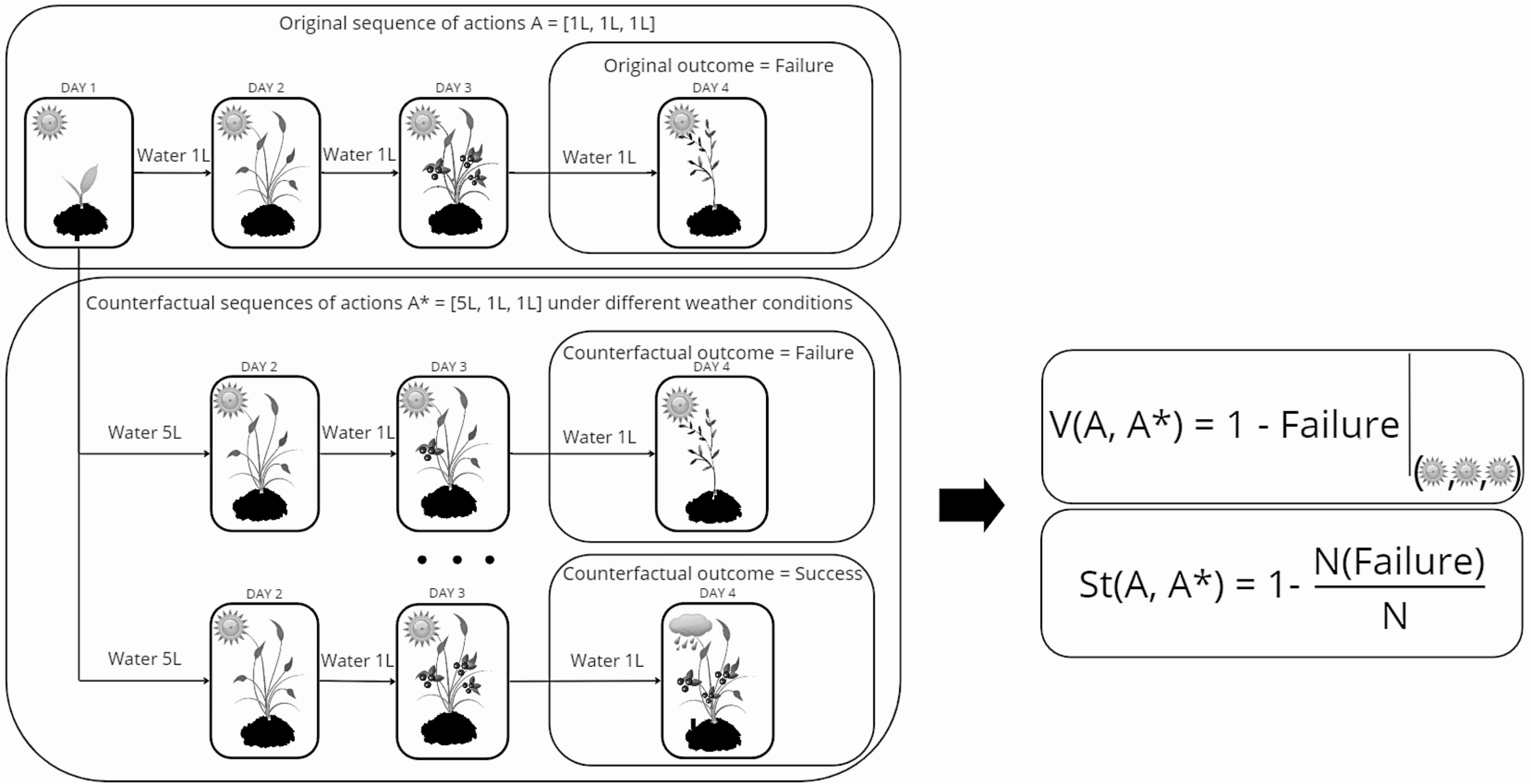}
    \caption{Stochastic uncertainty and validity evaluation in ACTER in a farming scenario: In an episode that ends in a plant dying, an alternative sequence of actions that increases water in the first action is examined. We run simulations of the episode with an alternative sequence of actions under different stochastic conditions represented by the weather. Validity indicates whether the plant would die after the alternative sequence of actions under the sunny weather conditions present in the original episode. Stochastic uncertainty is calculated as the probability of the plant surviving over all simulated weather conditions. }
    \label{stoch}
\end{figure}

\subsubsection{Proximity}

To minimize user effort in preventing failure we want to ensure that sequences A and A* are as similar as possible. 
For example, a counterfactual should not instruct the user to reduce their speed by a lot to avoid a crash when even a smaller speed reduction would have been enough. To that end, we define proximity between the original sequence $A$ and counterfactual sequence $A^*$ as:

\begin{equation}
    P(A, A^*) = 
    \begin{cases}
    
    \frac{\sum_{i=1}^{k} (a_{n-i} \neq a^*_{n-i})}{k} & \text{if A is discrete} \\
      \frac{\sum_{i=1}^{k} |a_{n-i} - a^*_{n-i}|}{k} & \text{if A is continuous} 
    \end{cases}   
\end{equation}

\subsubsection{Sparsity}

Research has shown that users prefer selective explanations, focusing only on a few most important elements \cite{miller2019explanation}. For this reason, we define sparsity as a metric that minimizes the number of actions that need to be changed to avoid failure. Given an original sequence of actions $A = [a_{n-k}, \dots, a_{n-1}]$ and a counterfactual sequence $A^*= [a^*_{n-k}, \dots, a^*_{n-1}]$, we define sparsity as:

\begin{equation}
    S(A, A^*) = \sum_{i=1}^{k} (a_{n-i} \ne a_{n-i})
\end{equation}

\subsubsection{Stochastic uncertainty}

With validity, we ensure that the counterfactual sequence $A^*$ successfully avoids failure in the specific episode. However, due to the stochasticity in the environment, the same sequence of actions might not avoid failure if the stochastic configuration was slightly different. Providing users with counterfactuals that are valid only for one stochastic configuration and might fail as a result of a slight change in the environment can lead the users to lose trust and discourage interaction with the system. Our goal is to provide counterfactuals that are robust to stochastic processes in the environment and can avoid the negative outcome with high certainty. For a counterfactual sequence of actions $A^*$ executed from state $x_{n-k}$, we define stochastic uncertainty as:

\begin{equation}
    S_t(A^*) = P[F(A^*(x_{n-k})|_{W}) = 0 \quad \forall W \in \mathcal{P}]
\end{equation}

where $\mathcal{P}$ is a set of all possible stochastic configurations. Intuitively, stochastic uncertainty ensures that $A^*$ avoids failure with high certainty regardless of stochastic processes in the environment, such as the behavior of other actors. A visualization of how stochastic uncertainty is calculated is shown in Figure \ref{stoch}.

\subsubsection{Recency}

When thinking about counterfactuals, humans prefer to change the most recent events to change the outcome \cite{byrne2019counterfactuals}. Telling the user that the car crash could have been avoided if agent never entered the street of the crash is not as useful as showing that a small reduction in speed a few moments earlier could have prevented failure. For this reason, we introduce recency as an important metric for diagnosing and correcting failure. For a sequence of action $A$ leading to failure and a counterfactual sequence $A^*$ we define recency as: 

\begin{equation}
    R(A, A^*) = \sum_{i=1}^{k} (a_{n-i} \ne a^*_{n-i}) * \frac{2}{k(k+1)}
\end{equation}

The coefficient $\frac{2}{k(k+1)}$ is chosen so that recency always takes values between 0 and 1.

\subsection{Multi-objective Counterfactual Search}
\label{Algorithm}

Our goal is to generate counterfactual sequences that optimize the defined counterfactual properties. However, some of these properties contradict each other and tradeoffs between them have to be made. Additionally, different users might deem different properties as more important. For example, while one user might only be interested in highly certain counterfactual sequences regardless of the number of actions that have to be changed, another might prefer changing fewer actions even if it comes with a greater risk of failure.  By receiving diverse counterfactual sequences, users can get a better understanding of the different ways to prevent failure and choose the one that best fits their preferences.

Previous work often conflates the counterfactual properties into a single objective function that can be optimized \cite{huber2023ganterfactual, olson2021counterfactual}. However, such approaches require fine-tuning of weights corresponding to different counterfactual properties. Additionally, these approaches only produce one counterfactual and have to be run multiple times with different sets of weights to obtain a set of diverse solutions for the counterfactual properties. In contrast, we use multiobjective optimization which simultaneously optimizes different objectives and generates a set of solutions. Specifically, given a sequence of actions $A = [a_{n-k}, \dots, a_{n-1}]$
failing, ACTER generates an alternative sequence of actions $A^* = [a_{n-k}^*, \dots, a_{n-1}^*]$ such that it minimizes the counterfactual properties defined in Section \ref{Counterfactual Properties}:

\begin{equation}
    \argmin_{A^*} [1 - V(A^*), P(A, A^*), S(A, A^*), 1 - S_t(A^*),  R(A, A^*)]
\end{equation}

\begin{algorithm}[t]
\caption{Non-dominated Sorting Genetic Algorithm (NSGA-II) -- Main loop}
    \begin{algorithmic}[1]
        \STATE \textbf{Input}: initial population $P_0$ \hfill \COMMENT{Counterfactual sequences sampled from the neighborhood of failure sequence}
        \STATE 
        \quad \quad \quad  constraint set $\mathcal{C} = [V(A, A^*)]$ \hfill \COMMENT{Set of constraints includes only validity}
        \STATE 
        \quad \quad \quad  objective set $\mathcal{O}= [P(A, A^*), S(A, A^*), S_t(A, A^*), R(A, A^*)]$ \hfill \COMMENT{Remaining counterfactual properties}
        \STATE \textbf{Parameters}: population size $N$, number of generations $G$
        \STATE $Q_0$ = generate\_child\_population($P_0$) \hfill \COMMENT{Generates child population using selection, recombination and mutation}
        \STATE $i = 0$
        \STATE $P_i, Q_i = P_0, Q_0$
        \WHILE{i < G}{
            \STATE $R_i = P_i + Q_i$ \hfill \COMMENT{Combine parent and child population}
            \STATE $\mathcal{F}_1, \dots, \mathcal{F}_t$ = fast\_nondominated\_sorting($R_i, \mathcal{C}, \mathcal{O}$) \hfill \COMMENT{Generate sets of nondominated solutions} 
            \STATE $P_{i+1} = \{\}$ \hfill \COMMENT{Initialize next parent population}
            \STATE $l = 1$
            \WHILE{|$P_{i+1}| + |F_{l}| < N $}{
                \STATE $P_{i+1} = P_{i+1} \cup \mathcal{F}_l$ \hfill \COMMENT{Fill $P_{i+1}$ with solutions starting with the best ones}
            }\ENDWHILE
            \IF{$|P_{i+1}| < N$}{
                \STATE $C_{dist}$ = calculate\_crowding\_distance($\mathcal{F}_i$) \hfill \COMMENT{Calculate crowding distance for the last nondominated set}
                \STATE $\mathcal{F}^*_{l}$ = sort\_descending($\mathcal{F}_l, C_{dist}$) \hfill \COMMENT{Sort descending based on the crowding distance}
                \FORALL{$f \in \mathcal{F}^*_{l}$}{
                    \STATE $P_{i+1} = P_{i+1} \cup f$ \hfill \COMMENT{Include the best solutions from the final front until size $N$ is reached}
                    \IF{$|P_{i+1}| >= N$}{
                        \PRINT
                    }\ENDIF
                }\ENDFOR
            } \ENDIF
        }\ENDWHILE
    \end{algorithmic}
    \label{nsga}
\end{algorithm}

We employ NSGA-II (Non-dominated Sorting Genetic Algorithm) \cite{deb2002fast}, an evolutionary algorithm that provides a set of non-dominated solutions on the Pareto front. This way we can generate a set of diverse counterfactuals that optimize different counterfactual objectives.
We use validity as a constraint for the algorithm and only consider solutions that satisfy it. The other four counterfactual properties -- proximity, sparsity, stochastic uncertainty, and recency are used as objectives in the algorithm and are simultaneously optimized. Additionally, we initialize the search with a dataset of sequences similar to the original one. We generate this dataset by applying noise to elements of the original sequence. This way search is initialized to the neighborhood of the initial sequence. NSGA-II modifies the initial population through generations while maintaining a front of the most promising solutions based on the constraints and objectives. In each iteration $i$, NSGA-II uses selection,  recombination and mutation to create a child population $Q_i$ from a parent population $P_{i}$. The parent and child populations are combined into $R_i = P_i \cup Q_i$ and filtered together to obtain the next parent population $P_{i+1}$. Firstly, a fast nondominated sorting algorithm is used to split $R_i$ into different fronts of nondominated solutions $\mathcal{F_1}, \dots, \mathcal{F}_t$, where $\mathcal{F}_1$ contains the best solutions and so on. $P_{t+1}$ is then populated using fronts starting with the best $\mathcal{F}_1$ until size $N$ is reached.  Finally, if $F_{l}$ is the last front that can be included, it is sorted according to the crowding distance. This ensures that in case of equally good solutions, the one from a less populated area of solution space is selected providing a diversity of solutions. NSGA-II algorithm, as applied to ACTER is described in more detail in Algorithm \ref{nsga}.

\section{Experimental Evaluation}
In this section, we introduce evaluation tasks (Section \ref{Evaluation Environments}) and baselines (Section \ref{Baselines}). In Section \ref{Evaluation Hypotheses} we define evaluation hypotheses and in Section \ref{Diversity Metrics} we define diversity metrics for sets of counterfactual explanations.

\subsection{Evaluation Environments}
\label{Evaluation Environments}

We evaluate ACTER in two complex, multi-objective and stochastic RL tasks -- highway driving and Farm environment.

\subsubsection{Highway Environment}

In the highway environment \cite{highway-env}, the agent is tasked with driving on a multi-lane highway. The agent can observe the coordinates and speed of neighboring vehicles and at each step choose one of five discrete actions -- change lane to the left, change lane to the right, speed up, slow down or do nothing (i.e., continue driving without changing the lane or speed). During training, agent has to optimize for multiple criteria such as avoiding collisions, maintaining good speed according to the speed limit, and remaining in the rightmost lane. Stochasticity in this environment comes from the behavior of other vehicles on the road which operate independently of the agent. As a failure in this environment, we choose the event of colliding with another vehicle. Alternatively, ACTER could be used to explain several different unwanted outcomes, such as changing a lane from the rightmost one, exhibiting too high or too slow speed or failing to maintain an appropriate distance from other vehicles. 

\subsubsection{Farm}

Farm-gym \cite{maillard2023farm} is a collection of environments that simulate diverse agricultural tasks. We evaluate ACTER in a simple Farm environment that involves growing a tomato plant. At each time step, the agent can choose to either apply between 1L and 10L of water or to harvest the plant. Positive rewards are obtained by harvesting a ripe plant, and every time a plant transitions to a different growth stage. Stochasticity of the environment is exhibited in the weather conditions, such as temperature, humidity, and rain. As a failure, we consider the event when the plant dies, which can happen after being watered too much or too little or after failure to harvest a ripe plant. However, ACTER could be used to explain various other outcomes of interest, such as applying too much water or harvesting before the optimal time. 

\begin{table}
\centering
    \begin{minipage}[b]{0.46\textwidth}\centering
    \caption{Parameters used for training a DQN black-box model $M$ in highway and Farm environments.}
    \begin{adjustbox}{width=\textwidth}
        \begin{tabular}{c|cc} \toprule 
            \backslashbox{Parameter}{Task} & Highway driving  & Farm \\ \midrule
            Number of layers & $2$ & $2$ \\
            Nodes in each layer & $[128, 128]$ & $[128, 128]$\\
            Learning rate & $5\cdot 10^{-4}$ & $9 \cdot 10^{-5}$ \\
            Training time steps ($N$) & $2 \cdot 10^5$ & $2 \cdot 10^4$\\
             \bottomrule    
    \end{tabular}
    \end{adjustbox}
    \label{params-train}
    \end{minipage}
    \hfill
    \begin{minipage}[b]{0.48\linewidth}\centering
        \caption{Parameters used for generating counterfactual explanations ACTER in highway and Farm environments.}
        \begin{adjustbox}{width=\textwidth}
            \begin{tabular}{c|cc} \toprule 
                \backslashbox{Parameter}{Task} & Highway driving  & Farm \\ \midrule
                Number of generations ($G$)  & $5$ & $10$ \\
                Population size ($N$) & $50$ & $100$\\
                Horizon ($k$) & $5$ & $10$\\
                Evaluation dataset size ($|D|$) & $247$ & $62$ \\
                 \bottomrule    
            \end{tabular}
        \end{adjustbox}
        \label{params}
    \end{minipage}
\end{table}

\subsection{Baselines}
\label{Baselines}

The goal of ACTER is to identify the changes in past decisions that could have helped agents prevent an undesirable outcome. To the best of our knowledge, ACTER is the first approach that can offer a counterfactual sequence of actions that prevents an unwanted outcome. Previous approaches in counterfactual state explanations can indicate which state features can be changed to illicit a different outcome, but cannot attribute failure to a past action, or provide an alternative action sequence to avoid failure \cite{olson2021counterfactual,huber2023ganterfactual}. For this reason, we compare ACTER to approaches that can examine an agent's path and attribute outcomes to previous actions, as follows:

\begin{enumerate}
    \item HIGHLIGHTS: based on the HIGHLIGHTS algorithm \cite{amir2018highlights} for detecting important moments in the agent's execution. HIGHLIGHTS identifies important states as those in which there is a substantial difference in outcomes when different actions are taken. Specifically, HIGHLIGHTS selects states that maximize the difference in estimated Q values between the most and least promising actions $max(Q(s, \mathcal{A})) - min(Q(s, \mathcal{A}))$.  
    \item CERTAIN: identifies important execution moments based on the agent's certainty when making a decision based on a metric proposed by \citet{sequeira2020interestingness}. A state is considered important if the agent is certain of its decision in the state. We measure the certainty using the entropy of Q values in state $s$. Intuitively, the agent could be failing because it is overly confident in certain incorrect behaviors, and changing the action the agent takes in these states could prevent failure.  
    \item UNCERTAIN: based on a metric opposite to the one used in CERTAIN \cite{sequeira2020interestingness}, it identifies important execution moments as states where the agent is uncertain. We calculate the uncertainty of a state as the entropy of the Q values. Intuitively, the agent could be failing because it is not fully trained and still uncertain of correct behavior in certain states. Changing which action the agent took in these uncertain states could prevent the failure.
    \item LOCAL\_MIN: based on the algorithm introduced by \citet{sequeira2020interestingness}, this baseline considers important the states in which the agent reaches a local minimum in terms of the reward function. Local minimum states could indicate why the agent failed, and preventing the agent from arriving to these states could prevent failure.
    \item LOCAL\_MAX: based on a metric opposed to LOCAL\_MIN, this algorithm considers a state important if it represents a local maximum in terms of the agent's state value. Local maxima states show states in which the agent considers itself to be in a favorable position. However, if an agent is not correct in its valuation of what a favorable position is, changing its action leading to a local maximum could prevent failure. For example, if an autonomous vehicle favors high speed, changing its decision that speed up the vehicle to a local maximum could be useful for a user who values maintaining a lower speed. 
\end{enumerate}

These baseline approaches can only detect important actions but cannot recommend alternative actions that could prevent failure. To enable a fair comparison, we extend these approaches to enable this additional functionality. Specifically, we try different random actions instead of the one selected by the algorithm as important and choose the ones that satisfy the validity constraint. In this way, we get an alternative sequence of actions that can avoid failure and can be compared to ACTER. To increase diversity we consider all sequences of actions that satisfy validity.

\subsection{Evaluation Hypotheses}
\label{Evaluation Hypotheses}

To evaluate ACTER we compare it to baselines described in Section \ref{Baselines} in highway driving and Farm environments. We establish four evaluation hypotheses for evaluating ACTER: 

\begin{itemize}
    \item \textbf{H1:} ACTER can produce counterfactual sequences that prevent failure with lower effort and higher certainty in stochastic environments compared to the baselines.
    \item \textbf{H2:} ACTER can produce a set of counterfactual sequences that offer more diverse ways of preventing failure compared to the baselines.
    \item \textbf{H3:} Actionable counterfactual sequences generated by ACTER help users better understand, diagnose, and correct failure compared to non-actionable explanations that only identify actions responsible for the failure.
    \item \textbf{H4:} Explanations generated by ACTER will
    be perceived as more satisfactory by users compared to non-actionable explanations that only identify actions responsible for the failure.
\end{itemize}

\begin{table*}[t]
    \centering
    \caption{The average values of counterfactual properties for counterfactual sequence generated using ACTER and the baseline approaches in highway driving and Farm environments.}
    \begin{adjustbox}{width=\linewidth}
    \begin{tabular}{c|cccccc|cccccc} \toprule 
        Task & \multicolumn{6}{c|}{Highway driving} & \multicolumn{6}{c}{Farm} \\ \midrule
        \backslashbox{Metric}{Approach} & HIGHLIGHTS & CERTAIN & UNCERTAIN & LOCAL\_MIN & LOCAL\_MAX & ACTER & HIGHLIGHTS & CERTAIN & UNCERTAIN & LOCAL\_MIN & LOCAL\_MAX & ACTER\\ \midrule
        Generated counterfactuals (\%) & 23 & 15 & 53 & 20 & 44 & \textbf{89} & 92 & 90 & 0 &0 & 21 & \textbf{100} \\ \midrule
        Validity ($\uparrow$) & \textbf{1.00} & \textbf{1.00} & \textbf{1.00} & \textbf{1.00} & \textbf{1.00} & \textbf{1.00} & \textbf{1.00} & \textbf{1.00} &  - & - & \textbf{1.00} & \textbf{1.00} \\ \midrule
        Proximity ($\downarrow$) & \textbf{0.20} & \textbf{0.20} & \textbf{0.20} &\textbf{ 0.20} & \textbf{0.20} & 0.30 & \textbf{0.10}& \textbf{0.10} & - & - & \textbf{0.10} & 0.14\\ \midrule
        Sparsity ($\downarrow$)  & \textbf{0.20} & \textbf{0.20} & 0.20 & \textbf{0.20} &\textbf{ 0.20} & 0.30 & \textbf{0.10} & \textbf{0.10} & - & - & \textbf{0.10} & 0.42 \\ \midrule
        Stochastic uncertainty ($\uparrow$)  & 0.66 & 0.62 & 0.75 & 0.61 & 0.76 & \textbf{0.85} & \textbf{0.0} & \textbf{0.0} & - & - &\textbf{0.0} & \textbf{0.0} \\ \midrule
        Recency ($\downarrow$) & 0.18 &\textbf{0.09} & 0.25 & 0.12 &  0.31 & 0.25 & 0.07 & 0.04 & - & - & \textbf{0.03} & 35 \\ 
        \bottomrule    
    \end{tabular}
    \end{adjustbox}
    \label{results}
\end{table*}

To evaluate H1, we evaluate the counterfactual properties of counterfactual sequences generated by ACTER and the baseline approaches. The results are presented in Section \ref{Evaluating Counterfactual Properties}. Hypothesis H2 is presented in more detail in Section \ref{Evaluating Diversity}. Hypotheses H3 and H4 are evaluated through a user study and the results are presented in Section \ref{User Study}.

\subsection{Diversity Metrics}
\label{Diversity Metrics}

While previous work offered metrics to estimate diversity for a set of counterfactual states \cite{mothilal2020explaining}, these do not apply to sequences of actions. For that reason, we introduce three metrics for evaluating the diversity of counterfactual sequences. Given a set of failure states $X_f$, a set of counterfactual actions $A^*(x^i)$ is generated for each $x^i \in X_f$. 

\begin{enumerate}
    \item \textit{Coverage (C):} measures the average number of generated counterfactual sequences per each failure:
    \begin{equation}
        C = \frac{1}{|X_f|}  \sum\limits_{x^i \in X_f} |A^*(x^i)|
    \end{equation}
    \item \textit{Action diversity (AD)}: measures the average distance between generated action sequences $A^*$:
    \begin{equation}
        AD = \frac{1}{|X_f|} \sum\limits_{x^i \in X_f}
        \sum\limits_{\substack{A^*_i, A^*_j \in A^*(x^i) \\  A^*_i \ne A^*_j}}
        Dist(A^*_i, A^*_j) 
    \end{equation}
    where the distance between two action sequences $Dist(A_i, A_j)$ is calculated as:
    \begin{equation}
        Dist(A_i, A_j) = \begin{cases}
             \frac{1}{k}\sum_{a \in A_i, b \in A_j} a \ne b & \text{if $\mathcal{A}$ is discrete} \\
             \frac{1}{k}\sum_{a \in A_i, b \in A_j} |a - b| & \text{if $\mathcal{A}$ is continuous}
        \end{cases}
    \end{equation}

    \item \textit{Counterfactual property diversity (CPD):} measures the average difference in counterfactual properties for generated counterfactual sequences $A^*$. Given a set of counterfactual properties $\mathcal{C}$: 
    \begin{equation}
        CDP(A^*) =  \frac{1}{|X_f|} \sum\limits_{x^i \in X_f} \sum\limits_{\substack{A^*_i, A^*_j \in A^*(x^i)\\ A^*_i \ne A^*_j}} CFDist(A^*_i, A^*_j) 
    \end{equation}

    where $CFDist(A_i, A_j)$ is defined as:
    \begin{equation}
        CFDist(A_i, A_j) = 
             \sum_{c \in \mathcal{C}} (c(A_i) - c(A_j))^2
    \end{equation}

\end{enumerate}

Coverage ensures that several options are offered to the users. Action diversity allows the user to choose an action sequence that best fits their preference. For example, if a car crash could have been prevented by either slowing down or changing lanes, a user who is in a rush might prefer to change lanes instead of sacrificing speed. Counterfactual property diversity ensures that the set of solutions exhibits different trade-offs of the counterfactual properties. For example, some users might prefer to change more actions but ensure that failure is prevented with higher certainty, while others might be more inclined to take a risk while changing fewer features.

\begin{table*}[t]
    \centering
    \caption{The diversity evaluation results for counterfactual sequences generated using ACTER and the baseline approaches in highway driving and farming environments.}
    \begin{adjustbox}{width=\linewidth}
    \begin{tabular}{c|cccccc|cccccc} \toprule 
        Task & \multicolumn{6}{c|}{Highway driving} & \multicolumn{6}{c}{Farm} \\ \midrule
        \backslashbox{Metric}{Approach} & HIGHLIGHTS & CERTAIN & UNCERTAIN & LOCAL\_MIN & LOCAL\_MAX & ACTER  & HIGHLIGHTS & CERTAIN & UNCERTAIN & LOCAL\_MIN & LOCAL\_MAX & ACTER \\ \midrule
        C ($\uparrow$) & 1.30   & 1.72 & 1.46 & 1.53 & 1.23 & \textbf{2.36} & 1.0 & 1.0 & 0.0 & 0.0 & 1.0 & \textbf{2.04}\\ \midrule
        AD ($\uparrow$) & 0.10  & 0.15 & 0.13 & 0.14 & 0.09 & \textbf{0.37} & 0.0 & 0.0 & 0.0 & 0.0 & 0.0 & \textbf{0.09} \\ \midrule
        CPD ($\uparrow$) & 0.06 & 0.08 & 0.04 & 0.13  & 0.08&  \textbf{0.19} & 0.0& 0.0& 0.0& 0.0& 0.0& \textbf{0.01}\\ 
        \bottomrule    
    \end{tabular}
    \end{adjustbox}
    \label{diversity}
\end{table*}

\section{Results}

We evaluate ACTER against five baselines in highway and Farm tasks. Firstly, we obtain a black-box policy $M$ by training a DQN for $T$ steps. The training parameters for policy $M$ are presented in Table \ref{params-train}. While this policy exhibits good behavior in the environment, due to its complexity there are still situations in which it can fail. In the highway task, we run $M$ for $1000$ and record failure in $247$ episodes. In the Farm task, we execute $M$ for $2000$ and record failure in $62$ episodes. For each of the failures, we use ACTER and baselines described in Section \ref{Baselines} to generate counterfactual sequences of actions that could prevent the failure. The parameters used by these approaches are shown in Table \ref{params}.

\subsection{Evaluating Counterfactual Properties}
\label{Evaluating Counterfactual Properties}

To evaluate H1 we evaluate ACTER and the baselines on counterfactual properties validity, proximity, sparsity, stochastic uncertainty, and recency, as defined in Section \ref{Counterfactual Properties}. For each of the examined algorithms, we generate counterfactuals for failure trajectories in highway and Farm environments. The results are presented in Table \ref{results}.

\textbf{ACTER successfully generates counterfactuals for most failures in both tasks and performs the best on stochastic uncertainty property, partially confirming H1.} In the  
highway environment ACTER finds a counterfactual sequence of actions that avoids failure in $89\%$ of trajectories. Among the baseline approaches, UNCERTAIN generated the most counterfactuals in the highway environment at $53\%$, while CERTAIN generated the least with only $15\%$. This indicates that actions responsible for failure in the highway environment are the ones the agent was not confident about. Similarly, in the Farm environment, ACTER generated a counterfactual for each failure. CERTAIN and HIGHLIGHTS were the most successful among the baselines, generating counterfactuals for over $90\%$ of failures, while UNCERTAIN and LOCAL\_MIN do not generate counterfactuals for any failures in Farm task. This indicates that in the Farm task, failure can be attributed to actions in which the agent is falsely confident. ACTER shows the best results in stochastic uncertainty in both environments, generating counterfactuals that are most robust to changes in the stochastic conditions of the environment. 

In both tasks, each counterfactual satisfies validity. This is due to ACTER and baselines seeing validity as a hard constraint and generating only counterfactuals that avoid failure. For proximity, sparsity, and recency metrics, ACTER creates counterfactuals with worse values compared to baselines. As baselines are limited to changing only one action, they always generate counterfactuals with favorable proximity and sparsity values. ACTER, however, generates multiple solutions that optimize different metrics, leading to worse average proximity, sparsity, and recency. This corresponds to similar results in supervised learning where the trade-off between diversity and proximity has been shown \cite{mothilal2020explaining}. 

\subsection{Evaluating Diversity}
\label{Evaluating Diversity}

To evaluate hypothesis H2, we compare the counterfactuals generated by ACTER to baselines on the three diversity metrics defined in Section \ref{Diversity Metrics}. The results are presented in Table \ref{diversity}. 
 
\textbf{In both environments, ACTER performs better according to all three metrics, confirming H2.} It is important to note that ACTER was not trained to maximize any of the diversity metrics. In fact, ACTER generates diverse counterfactuals by design,  
due to its multi-objective optimization which simultaneously minimizes counterfactual properties generating a Pareto set of solutions. 

In the highway environment, ACTER generates on average $2.36$ counterfactual sequences. ACTER also achieves almost double values for action diversity compared to other approaches. Similarly, ACTER provides counterfactuals with the highest counterfactual property diversity scores. 

In the Farm task, all baselines generate no more than one counterfactual per failure. In contrast, ACTER generates an average of $2.04$ counterfactual sequences per failure. On the other two metrics, both ACTER and the baselines show low values. This is because, in this task, the plant often dies because it was not harvested in time. Failure is often prevented simply by executing a harvest action sooner in the episode. For this reason, there is little variety in explored counterfactual scenarios, resulting in lower values of action and counterfactual property diversity metrics.

\subsection{User Study}
\label{User Study}

Ultimately, counterfactual explanations are developed to help humans understand and engage with RL systems. While previous work on counterfactual state explanations has explored how users engage with state-based explanations, no user studies have explored how users perceive counterfactual explanations that examine and change previous actions. In this work, we make the first steps on the path of using counterfactual sequences with end-users. The goal of the study is to evaluate how explanations help users identify and correct failures as well as how they are perceived by the users. We compare explanations generated by ACTER to \textit{non-actionable explanations} that can only identify the actions responsible for an outcome but do not offer alternative actions that could change the outcome. Non-actionable explanations relay only a portion of the information that counterfactuals do but might ease the cognitive load of users. Previous work has found that in supervised learning counterfactual explanations can lead to an increase in false confidence compared to non-actionable explanations \cite{celar2023people}. The goal of this study is to examine this issue in RL. We obtain non-actionable explanations by running ACTER but showing
which action in the past can be changed while omitting which alternative action could prevent failure. We conducted the study in the highway task and showed participants examples of car crashes and explanations for this outcome. The link to the study is available at: \href{https://qrxhyre44mt.typeform.com/to/B2CvHr6a}{https://qrxhyre44mt.typeform.com/to/B2CvHr6a}.

We sourced 60 participants from English-speaking countries (UK, USA, Australia, New Zealand, Canada, and Ireland) through the Prolific platform and divided them into two groups. The first group received explanations generated by ACTER and the other non-actionable explanations. Participants were remunerated for their time according to the Prolific payment policy. The consisted of 3 parts -- training, testing, and subjective evaluation. Before the study, the participants were introduced to the task and given information about a specific type of explanation they would see during the study. In the training part, participants were shown examples of car crashes along with appropriate explanations. Each participant was presented with ten training examples. In the testing phase, participants were then shown examples of crashes without explanations and were asked to identify which previous action could be changed and how to prevent failure. Participants were shown 10 test questions. Due to visualization errors, one question had to be removed from the study, resulting in 9 test questions. The questions were selected such that there is only one alternative action sequence that would prevent failure. Finally, the last part of the study involved users rating explanations on a 1-5 Likert scale based on the \textit{explanation goodness metrics} \cite{hoffman2018metrics}. The metrics measured how satisfactory, useful, detailed, complete, actionable, trustworthy, and reliable users found the explanations. Users were also asked whether explanations made them more confident in their understanding of the agent's behavior similar to the work of \cite{celar2023people}. 

The participants to whom explanations generated by ACTER were given could identify the correct action in $42.59\%$, while those who have seen non-actionable explanations succeeded in $42.22\%$ of test cases. Furthermore, the former group successfully corrected the failure in $17.03\%$ and the latter group for $18.5\%$ of cases. This indicates that \textbf{there is no significant difference in users' ability to identify and correct failure between explanations generated by ACTER and non-actionable explanations}. These results reject hypothesis H3. 

On the other hand, users perceived explanations generated by ACTER as better on all explanation goodness metrics. We performed a non-parametric one-tailed Mann-Whitney U test and found that \textbf{participants perceived counterfactual explanations generated by ACTER to be significantly more useful ($ 0.0348$), detailed ($0.0085$) and actionable ($0.0465$)} compared to the non-actionable explanations. Additionally, \textbf{participants reported significantly higher confidence after seeing explanations generated by ACTER ($p = 0.0342$)}. However, no significant difference was detected between participants' ratings for satisfaction ($p = 0.0502$), completeness ($p = 0.2478$), reliability ($p =0.0747$) and trustworthiness ($p = 0.1115$) of explanations. As a result, H4 is only partially confirmed by our experiments.

Our study finds that counterfactual explanations do not help non-expert users better diagnose and correct failure. One reason for this could be the complexity of the task. Previous work on counterfactual state explanations focuses on state-based explanations and shows users one state and its explanation at a time \cite{olson2021counterfactual,huber2023ganterfactual}. Our study, however, required reasoning about longer chains of events and learning connections between multiple states and actions and the outcome. We speculate that counterfactuals on their own might not be enough to fully explain the behavior of an agent in such a complex scenario to non-expert users. In future work we hope to explore if these results hold for experts that are familiar with RL systems and if counterfactuals can help them in debugging suboptimal policies. Additionally, we hope to explore how counterfactuals can be combined with other types of explanations to be best applied to explain policies to non-expert users.

The outcomes of our study correspond to similar work in supervised learning where counterfactuals have been found to provide false confidence when predicting outcomes \cite{celar2023people}. That counterfactuals can lead to false confidence in RL is a worrying discovery as it can lead users to confidently engage with the RL system without truly understanding it. On the other hand, counterfactuals are perceived as more detailed, useful, and actionable, indicating the potentially wider range of information that they convey. Further research is needed to understand how counterfactuals can be best utilized in RL.

\section{Future Work}
\label{Future Work}

In this work, we proposed ACTER, an approach for generating actionable and diverse counterfactuals for diagnosing and correcting outcomes in RL. We defined and implemented five counterfactual properties and provided an evolutionary algorithm for the multi-objective optimization. Moreover, we proposed three diversity metrics to benchmark the diversity of counterfactual explanations. Finally, we conducted a user study and showed that while users prefer counterfactual explanations, they are not more helpful in learning to diagnose and correct failure than non-actionable explanations. 

In future work, we hope to use encoders to better approximate proximity in discrete action spaces. Additionally, in this work, we use simulations to estimate the stochastic uncertainty of an action sequence. This is only suitable for online scenarios where access to the environment is granted. In future work, we hope to explore how uncertainty can be estimated even from offline data. Finally, in this work we have shown that despite being preferred by the users, counterfactual explanations do not help them better diagnose and correct failure compare to simpler explanation types. In future work we plan to examine further the scenarios in which counterfactual explanations can be used both for explanation purposes as well as policy improvement and whether they would be more useful for expert users.


\section*{Acknowledgements}

This publication has emanated from research supported in part
by grants from Science Foundation Ireland under grant number
18/CRT/6223 and SFI Frontiers for the Future grant number 21/FFP-A/8957. For the purpose of Open Access, the author has applied a CC BY public copyright licence to any Author Accepted
Manuscript version arising from this submission.

\bibliographystyle{ACM-Reference-Format}
\bibliography{sample-base}

\newpage
\section*{Appendix}

\section{User Study Details}

\begin{figure}
    \centering
    \includegraphics[width=0.5\linewidth]{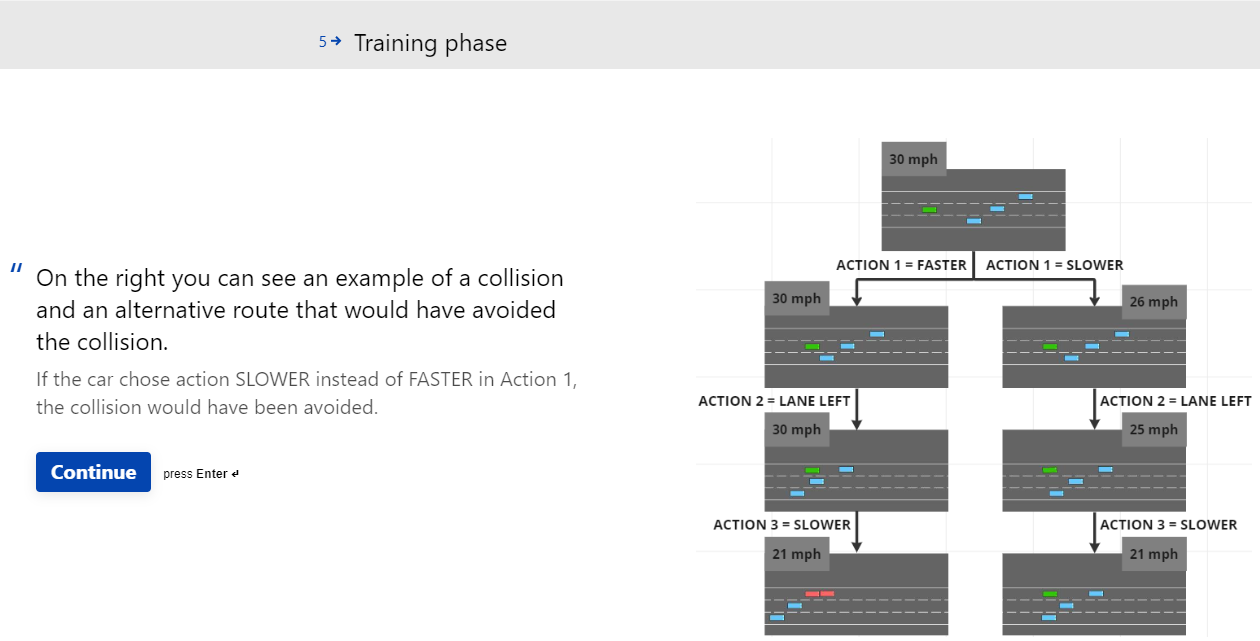}
    \includegraphics[width=0.42\linewidth]{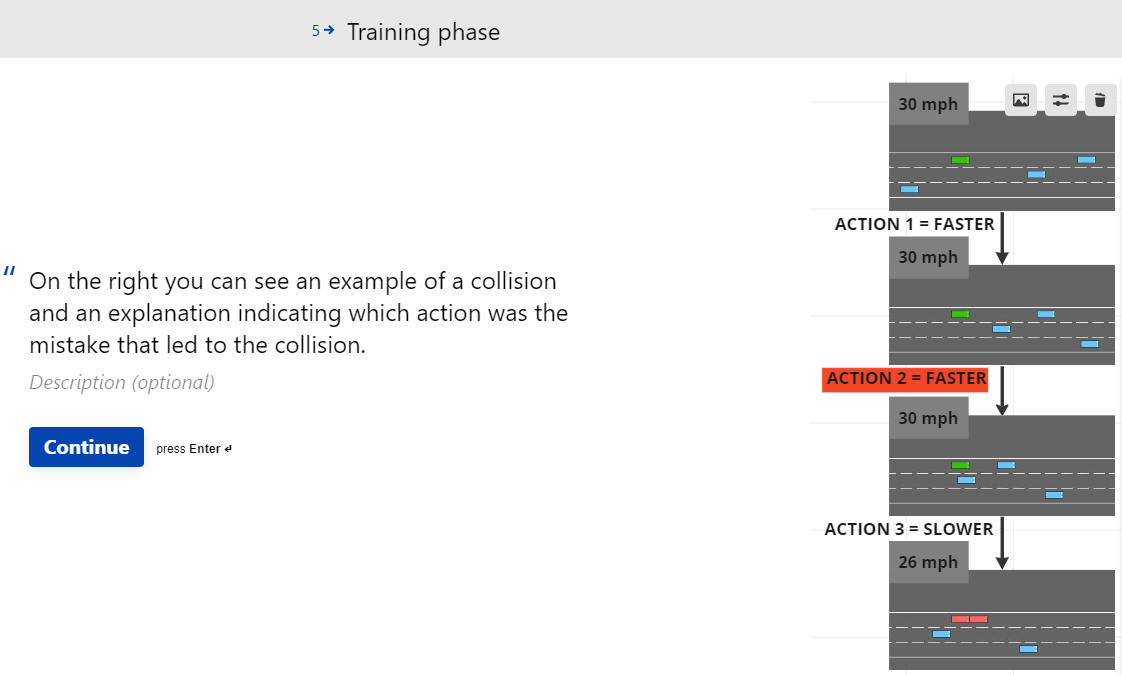}
    \caption{Examples of user study explanations shown to the users. Left: Counterfactual explanation. Right: Non-actionable explanation.}
    \label{example_e}
\end{figure}

\begin{figure}[t]
    \centering
    \includegraphics[width=0.45\linewidth]{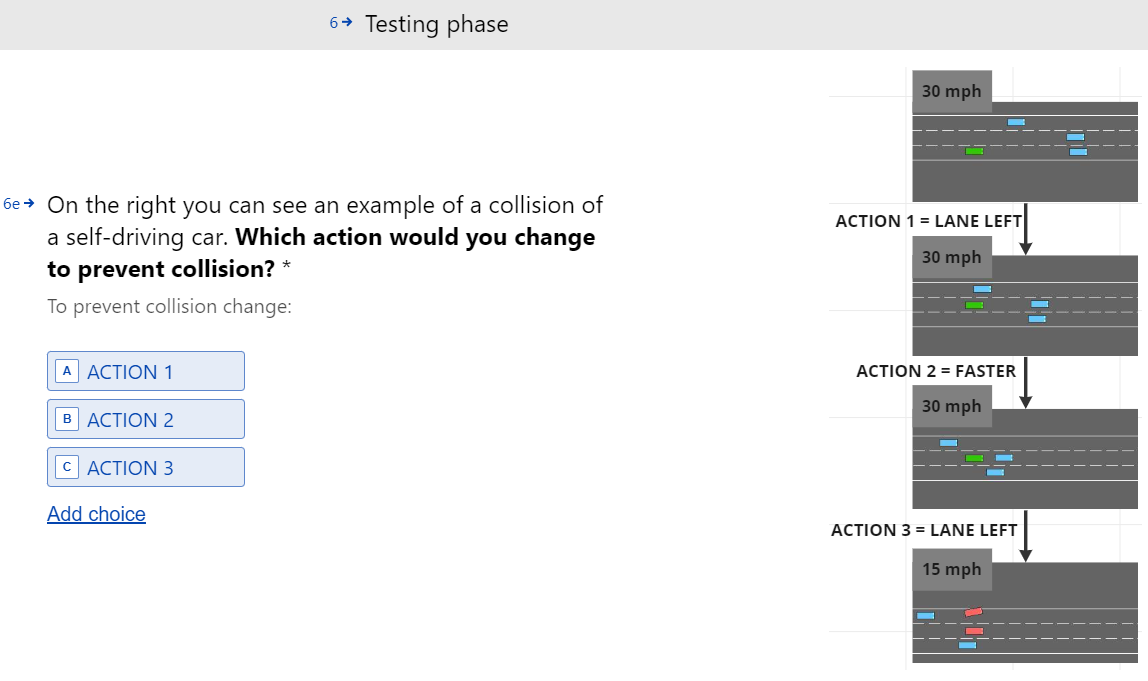}
    \includegraphics[width=0.45\linewidth]{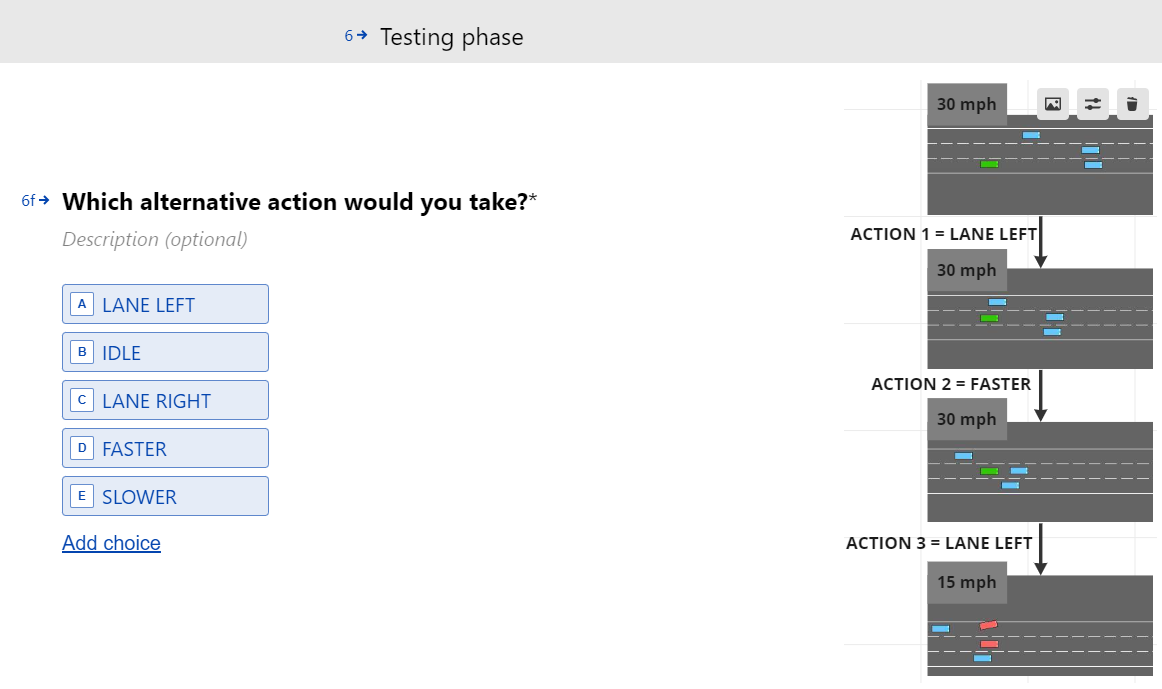}
    \caption{Examples of user study questions. Left: Question about identifying action responsible for failure. Right: Question about alternative action to prevent failure.}
    \label{example_q}
\end{figure}

In this section, we provide some examples of user study questions and templates used in this work. In Figure \ref{example_e} we show examples of a counterfactual explanation and a non-actionable explanation of a car crash presented to the users during the training phase of the study. Figure \ref{example_q} shows an example of a test question that the users were asked.

\end{document}